\documentclass[manuscript,screen]{acmart}
\usepackage{subfigure}
\AtBeginDocument{%
  \providecommand\BibTeX{{%
    \normalfont B\kern-0.5em{\scshape i\kern-0.25em b}\kern-0.8em\TeX}}}

\setcopyright{acmcopyright}
\copyrightyear{2023}
\acmYear{2023}
\acmDOI{10.1145/3575803}

\usepackage{graphicx}
\usepackage{multirow}

\usepackage{url}

\usepackage{booktabs}
\usepackage{caption}
\usepackage{float}

\usepackage{epsfig}
\usepackage{amsmath}

\usepackage{amssymb}

\usepackage{mathrsfs}
\usepackage{diagbox}
\usepackage{hyperref}

\usepackage{algorithm}
\usepackage{algorithmicx}
\usepackage{algpseudocode}

\def\eg{\emph{e.g.}} 
\def\ie{\emph{i.e.}}

\begin{document}
\title{Building Dialogue Understanding Models for Low-resource Language Indonesian from Scratch}

\author{Donglin Di}
\email{donglin.ddl@gmail.com}
\authornotemark[1]
\affiliation{
  \institution{Advance.AI}
}

\author{Weinan Zhang}
\email{wnzhang@ir.hit.edu.cn}
\affiliation{%
  \institution{Harbin Institute of Technology}
}

\author{Yue Zhang}
\email{yue.zhang@wias.org.cn}
\affiliation{
  \institution{Westlake University}
}

\author{Fanglin Wang}
\email{fanglin.wang@advance.ai}
\affiliation{%
  \institution{Advance.AI}
}



\begin{abstract}
Making use of off-the-shelf resources of resource-rich languages to transfer knowledge for low-resource languages raises much attention recently.
The requirements of enabling the model to reach the reliable performance lack well guided, such as the scale of required annotated data or the effective framework.
To investigate the first question, we empirically investigate the cost-effectiveness of several methods to train the intent classification and slot-filling models for Indonesia (ID) from scratch by utilizing the English data.
Confronting the second challenge, we propose a Bi-Confidence-Frequency Cross-Lingual transfer framework (BiCF), composed by ``BiCF Mixing'', ``Latent Space Refinement'' and ``Joint Decoder'', respectively, to tackle the obstacle of lacking low-resource language dialogue data.
Extensive experiments demonstrate our framework performs reliably and cost-efficiently on different scales of manually annotated Indonesian data.
We release a large-scale fine-labeled dialogue dataset (ID-WOZ) and ID-BERT of Indonesian for further research.
\end{abstract}



\maketitle

\section{Introduction}
\begin{figure}
    \centering
    \includegraphics[width=0.5\linewidth]{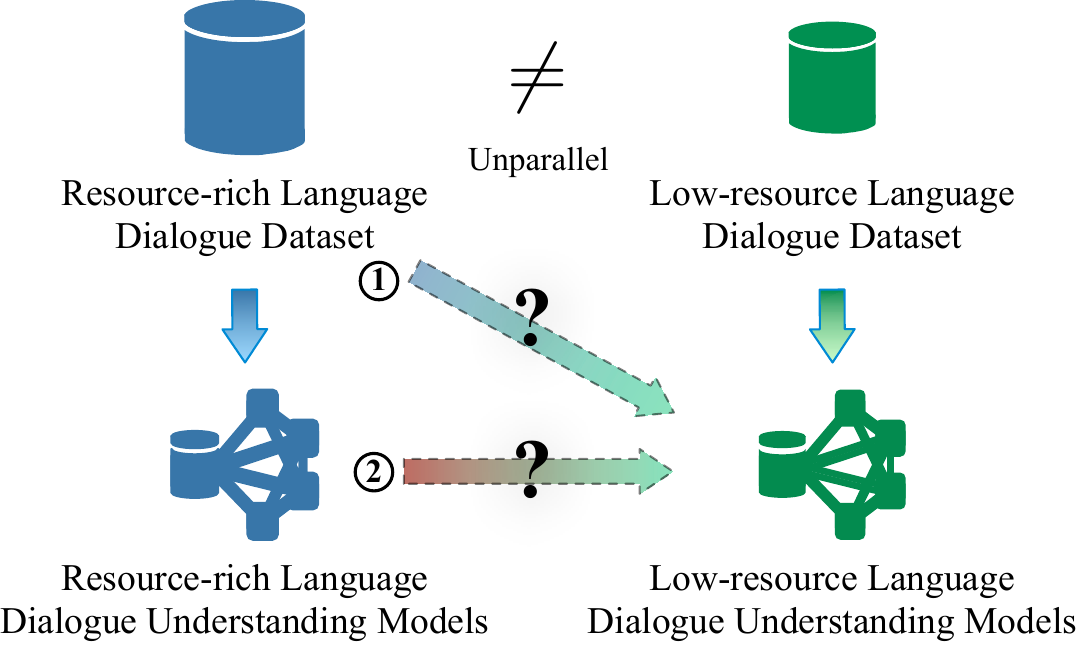}
    \caption{\label{fig:prob_eg} The investigation of utilizing off-the-shelf resources and models for generating low-resource language dialogue understanding models from scratch.}
\end{figure}

It is generally accepted that the neural dialogue understanding model relies heavily on large scale of training data \cite{liu2018towards}.
Existing works have been conducted mostly on rich-resource languages such as English\cite{wen2016network,lowe2015ubuntu} and Chinese\cite{wu2016sequential}.
But thousands of low-resource languages in this world lacks training data.
It is impractical and cost-ineffective to collect and annotate enough large-scale datasets for low-resource languages\cite{grave2018learning} to train the dialogue understanding models.
Therefore, as shown in Fig.~\ref{fig:prob_eg}, it remains huge challenges on how to efficiently adapt existing research resources and findings to low-resource languages (\eg, Indonesian (ID)), so that the needs for understanding the multilingual task-oriented dialogue\cite{schuster2018cross, schuster2019cross} can be addressed effectively.


One challenge is how to make use of off-the-shelf resources of rich-resource languages effectively (\eg, English \cite{budzianowski2018multiwoz}).
The intuitive method is to use a neural machine translation system \cite{vaswani2018tensor2tensor,cheng2019semi} to translate the English dataset into Indonesian, and then train the dialogue understanding models on the translated data.
Another strategy is to utilize multilingual word embeddings \cite{devlin2018bert, pires2019multilingual}, which allows the dialogue model trained on the English dataset to be directly applied to Indonesian since the pre-trained multilingual model contains vocabulary of both them.
Each of the above methods has its own strengths and limitations.
The former allows us to use language-specific pre-trained model \cite{peters2018deep, devlin2018bert, yang2019xlnet} for better representation, which could be obtained from training on large-scale unlabelled general documents or web pages.
However, methods of this branch suffer from errors in machine translation and invalid dislocated annotations from source corpus, which could significantly influence the subsequent dialogue modeling (\ie, slot-filling).
The latter suffers from intrinsic differences between English and Indonesian, including variations in syntactic and semantic patterns.

The second challenge relies on how to transfer existing models to the target low-resource language.
Cross-lingual transfer learning \cite{schuster2018cross} can potentially alleviate the limitations of both methods.
One possible approach is to align the contextual word embeddings or sentence-level encoding in the semantic latent space \cite{schuster2019cross}.
It can avoid semantic misunderstanding and syntactic mistakes.
However, this method can be impacted by imperfect alignments and implementation of it is complex, which leads to applying or deploying slowly.

In this work, we propose a \textbf{Bi}-\textbf{c}onfidence-\textbf{f}requency Cross-lingual Transfer framework (BiCF), overcoming challenges illustrated above. 
For the first challenge, we adopt the word-level alignment strategy \cite{zhang2019cross}, which has been demonstrated as effective as phrase-level alignment yet much simpler and more stable \cite{tiedemann2015improving, tiedemann2016synthetic}.
Specifically, the first stage of BiCF is Bi-confidence-frequency Mixing, utilizing the English dataset to generate code-mixed data, which avoids sentence-level translation errors as well as labels dislocation.
The mixed data not only takes the importance-frequency and translating-confidence into consideration but also contains gold annotations for Indonesian from English datasets.
And for the second challenge, in our framework, Latent Space Refinement and Joint Decoder are designed on the top of resulting high-quality mixed data, utilizing and refining the pre-trained off-the-shelf word embedding models, to train the dialogue understanding models (\ie, intent classification and slot-filling) for Indonesian eventually.

To conduct extensive experiments for Indonesian, we follow the method of MultiWOZ \cite{budzianowski2018multiwoz}, which is a large-scale task-oriented English dialogue dataset, to collect and annotate a counterpart and richer-domain dataset in Indonesian (ID), named ID-WOZ.
Extensive experiments show that the proposed framework can achieve comparable and better performance with less gold annotated data.
We quantify the influence of each factor in our experiments.

The main contributions of this paper are summarized as follows:
\begin{itemize}
    \item 
    We propose a framework (BiCF) to utilize English dataset for training Indonesian dialogue understanding models and achieve good performance on Indonesian dialogue dataset.
    \item 
    We release a large-scale manually annotated multi-domain ID-WOZ dialogue dataset, together with a pre-trained ID-BERT model as the resource contributions for low-resource language dialogue understanding tasks.
    \item
    We investigate the demand for annotated data of well-performing dialogue understanding models, which thereby may instruct related research works on collecting datasets or training models for other low-resource languages.
\end{itemize}
\section{Related Work}

\textbf{Low-resource Language.}
Exisiting works \cite{guo2015cross, duong2015neural, ammar2016many}, and \cite{wang2017universal} work on multilingual parsing on low-resource languages, which are helpful to improve the low-resource language understanding.
Wang et al. \cite{wang2017universal} propose to integrate English syntactic knowledge into a parser trained on the Singlish treebank, and shows that it is reasonable to leverage English to improve low-resource language models.\\
\textbf{Cross-lingual Transfer.}
Artetxe et al. \cite{artetxe2017learning} proposes a self-learning framework and a small size of word dictionary to learn a mapping between source and target word embeddings.
Zhang et al. \cite{zhang2019cross} focuses on improving dependency parsing by mixing confident target words into source treebank.
Schuster et al. \cite{schuster2019cross} utilizes Multilingual CoVe embeddings obtained from Machine Translation systems \cite{mccann2017learned} in Thai and Spanish for zero-shot dependency parsing.
Relying on aligned parallel sentence pairs suffers from noise and imperfect alignments \cite{zhang2019cross}.
In line with these methods, encoding the semantic information directly within the same cross-lingual latent space could avoid semantic misunderstanding from machine translation or wrong alignments.

\section{ID-WOZ}
\label{sec:dataset}
There has been a lack of available datasets for training natural dialogue understanding systems in regional low-resource languages, such as Indonesian \cite{chowanda2017recurrent,koto2016publicly,tho2018forming}.
As a result, we build our ID-WOZ dataset from scratch.
The detailed statistics is reported in Experiment Section.
In particular, we organize a structured annotation scheme with structured semantic labels, drawing on the experiences of previous work \cite{williams2013dialog, asri2017frames, eric2017key, shah2018building, budzianowski2018multiwoz}.

\subsection{Coverage}
ID-WOZ is constructed with the goal of obtaining highly natural conversations between a customer and an agent or a query information center focusing on daily life.
We consider various possible dialogue scenarios ranging from basic requests like \textit{hotel}, \textit{restaurant}, to a few urgent situations such as \textit{hospital} or \textit{police}.
Our dataset consists of nine domains, namely \textit{plane, taxi, wear, restaurant, movie, hotel, attraction, hospital, police}, most of which are extended domains including the sub-task \textit{Booking} (with the exception of \textit{police}).

\subsection{Collection and Annotation}
We adopt the Wizard-of-Oz \cite{kelley1984iterative} dialogue-collecting approach, which has been shown effective for obtaining a high-quality corpus at relatively low costs and with a small-time effort.
Following the successful experience of MultiWOZ \cite{budzianowski2018multiwoz}, we create a large-scale corpus of natural human-human conversations on a similar scale.
Based on the given templates for several domains, the users and wizards generate conversations using heuristic-based rules to prevent the overflow of information.
We design and develop a collection-annotation pipeline platform with a user-friendly structure for building up the dataset, shown in Experiment Section.

In order to accelerate and optimize the process of collection and annotating, we design and develop a pipeline platform.
Our platform consists of three stages, ``collection - annotation - statistics \& analysis'', which are executed synchronously after the initialization process.
We split a number of well-trained annotators (\ie, 80 local people, 70 of them whose mother language are ID, 10 of them are bilingual citizens, plus 2 main organizers) into two groups to produce dialogue and annotation.
A quarter of annotators (\ie, 20) are trained following the templates we provide to play the wizard role.
After collecting 1k dialogues initially (about one week), while the collecting conversation is still ongoing, the second group of annotators (\ie, 62) join in to work towards the detailed full-labeled corpus, including domains, actions, intents, and slots.

The quality is assured in three processes, namely ``scripts checking'', ``cross-checking'', and ``supervisor-checking''.
In particular, the scripts can filter the hypothesis cases which have potential faults such as vacant labels or malformed.
For cross-checking process, the annotators are assigned not only to fresh unlabeled annotation tasks but also a few sampling labeled cases (\ie, 20\%) from their peers, in a double-blind process.
The cases passing the cross-checking procedure will be sampled and handed over to the supervisors (\ie, the two organizers), who are more familiar with the details of the entire annotation task to control the overall consistency and accuracy of the annotation.
We adopt the inter-annotator agreement (IAA) \cite{fleiss1969large} to measure how well our recruited annotators can make the same annotation decision for a certain category further, as follows:
\begin{equation}
    \kappa \equiv \frac{p_{o}-p_{e}}{1-p_{e}}=1-\frac{1-p_{o}}{1-p_{e}}
\end{equation}
where $p_{o}$ and $p_{e}$ denote the relative observed agreement among raters, and hypothetical probability of chance agreement, respectively.
The average score of our dataset is 0.834.

\subsection{Statistics and Analysis}
Table~\ref{tab:dataset} compares our dataset with existing datasets in Indonesian, and the English dialogue dataset MultiWOZ \cite{budzianowski2018multiwoz}.
ID Chat \cite{koto2016publicly} is the first publicly available Indonesian chat corpora, and draw a few related research on the Indonesian Language dialogue \cite{chowanda2017recurrent}.
Dyadic Chat \cite{tho2018forming} is another public chat corpus on Indonesian, which focuses on the dyadic term.
Dyadic is a term that describes the relationship between two people, for instance, the romantic relationship between a couple.
Compared with these small-scale datasets, ID-WOZ is the first to contain large-scale (about ten thousand dialogues in multi-domains) corpus focusing on general task-oriented chat.

MultiWOZ \cite{budzianowski2018multiwoz} is a large-scale multi-domain task-oriented English dialogue dataset, including seven distinct domains (\textit{taxi, restaurant, hotel, attraction, hospital, police, train}) and fine-labeled actions and slots in the spoken language understanding stage.
Considering the regional cultural background, our collected dataset contains a few more general domains (\ie, \textit{wear, movie, plane}) and more corresponding slots type, such as \textit{clothes type, movie genre, movie synopsis}.

\begin{table}[t]
\begin{tabular}{l|rrrr}
\hline
\textbf{\textbf{Dataset}} & \textbf{\textbf{ID Chat}} & \textbf{\textbf{Dyadic Chat}} & \textbf{\textbf{MultiWOZ}} & \textbf{ID-WOZ} \\ \hline
Domains                   & None              & None                  & 7                & 9        \\
Language                  & ID                    & ID                        & En                    & ID       \\
Total \# dials            & 300                       & 79                            & 8, 438                        & 9, 189          \\
Total \# tokens            & 150, 000                      & 3164                          & 1, 520, 970                           & 1, 551, 591                   \\
Total \# utters           & 1, 000                        & 158                           & 142, 974                           & 251, 184                   \\
Avg. \# turns             & 3                         & 3                             & 13.68                      & 13.67              \\
Avg. \# slots             & -                         & -                             & 25                         & 8.8            \\ \hline
\end{tabular}
\caption{Comparison of ID-WOZ with other related datasets in several statistics metrics.}
\label{tab:dataset}
\end{table}

\section{ID-BERT}
Even though there are several off-the-shelf pre-trained BERT models for rich-resource languages such as English and Chinese, pre-trained language-specific models for low-resource languages like Indonesian, is still not available to our knowledge.
We release a pre-trained model for Indonesian named ID-BERT as another resource contribution.
Although most related work \cite{schuster2019cross, pires2019multilingual} relies on the pre-trained Multilingual-BERT model and fine-tunes it for low-resource languages, our main goal is to build up a relatively reliable dialogue system and to experiment how to overcome the gap between language-specific BERT and the multilingual-BERT.
Therefore, we put in effort to train a ID-BERT for comparing appearance on spoken language understanding or furthermore tasks.
We pre-train a BERT for Indonesian from scratch by about 3.3 billion tokens from Indonesian websites document-level corpus, which covers news reports, research papers, daily articles, and other text genres.
The size of our ID-BERT vocabulary is 0.9M, which is much larger than Multilingual-BERT (0.12M).
We believe that this size of the vocabulary is sufficient to cover most of the scenarios of daily multi-domain task-oriented dialogue in Indonesian.
The training takes one week by using Google Cloud TPU v3\_8, and our ID-BERT (Cased, L=12, H=768, A=12) are eventually obtained.

\section{BiCF Cross-lingual Transfer}
\label{sec:cross-lingual}

\begin{figure*}
    \centering
    \includegraphics[width=\linewidth]{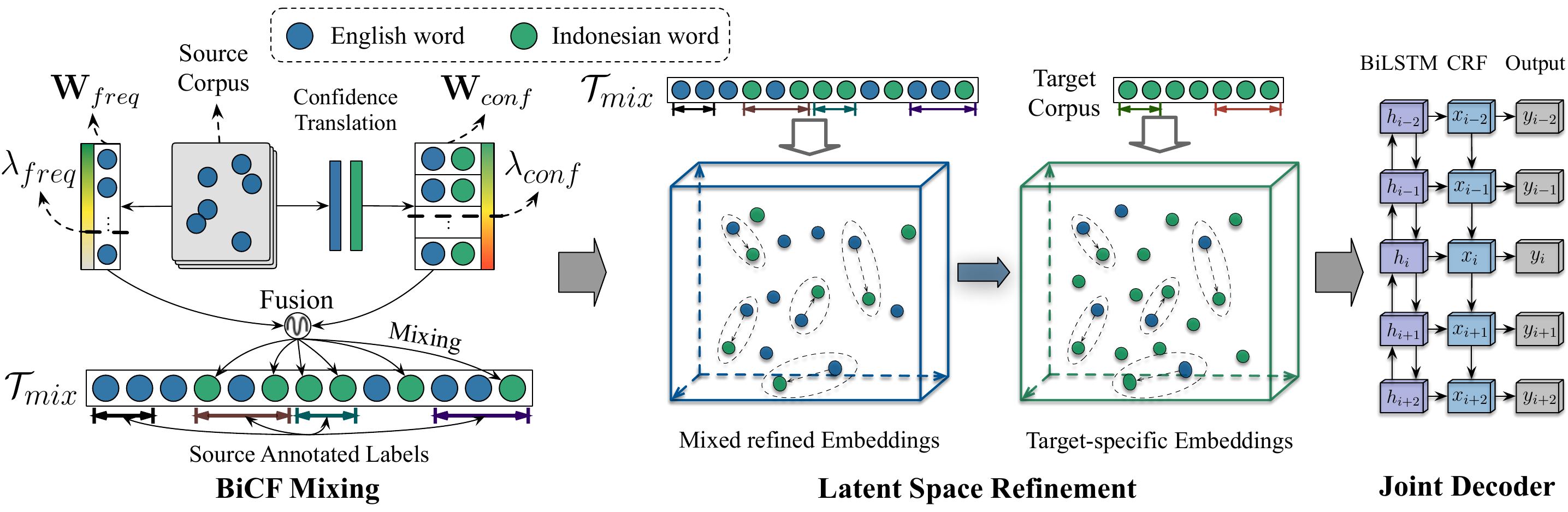}
    \caption{\label{fig:framework} Illustration of the proposed framework (BiCF), which consists of BiCF Mixing, Latent Space Refinement, and Joint Decoder. The frequency-word and confidence-word set in the first stage are derived from English dataset and confidence-translated parallel sentences, respectively. By fusion and mixing, the mixed data is generated. The cross-lingual space refinement module will generate a target-specific embedding model to represent Indonesian better. The final stage is to decode and output intent and slots jointly.}
\end{figure*}

We here illustrate our proposed pipeline framework -- ``BiCF Cross-lingual Transfer'' (BiCF) in detail.
It mainly consists of three components, namely ``BiCF Mixing'', ``Latent Space Refinement'', and ``Joint Decoder''.
As shown in Fig.~\ref{fig:framework}, the BiCF mixing step replaces a few English words with Indonesian.
Then we train and refine the cross-lingual semantic embedding latent space based on the mixed data with gold annotations from English dataset.
Finally, we adopt the combination of BiLSTM and CRF to decode the intent and slots jointly.

\subsection{BiCF Mixing}
The first stage of our framework is ``\textbf{Bi}-\textbf{c}onfidence-\textbf{f}requency Mixing'' (BiCF Mixing).
As shown in Fig.~\ref{fig:framework}, we use the source language data in two steps.
The first is to generate the frequency-word set ($\textbf{W}_{freq}$) of source data.
The second is to obtain the word alignment with the translating-confidence ($\lambda_{conf}$) of each word and generate confidence-word set ($\textbf{W}_{conf}$).
The goal of this stage is to select both frequent and high-confidence word pairs for English and Indonesian, and yield mixed data $\mathcal{T}_{mix}$.

Given the set of source sentences $\mathcal{S} = \{s_1, s_2, ..., s_n\}$, we calculate TF-IDF \cite{salton1982extended, ramos2003using} for each word in the source dialogue corpus, as shown in Eq.~\ref{tfidf}:
\begin{equation}
\label{tfidf}
\begin{aligned}
\left\{\begin{matrix}
tf_{(i,j)} &= \frac{\mathcal{N}(w_i^s, s_j)}{\sum_k \mathcal{N}(w_k^s, s_j)} \\
idf_{(i)} &= \log\frac{\left | S \right |}{1 + \left | j: w_i^s \in d_j \right |} \\
tf\texttt{-}idf_{(i,j)} &= tf_{(i,j)} \times idf_{(i)} \\
\end{matrix}\right.
\end{aligned}
\end{equation}
where $\mathcal{N}(w_i^s, s_j)$ is the number of occurrences of the word $w_i^s$ in the source language sentence $s_j$, and the denominator ($\sum_k \mathcal{N}(w_k^s, s_j)$) is the sum of occurrences of all terms $w_k^s$ in the sentence $s_j \in \mathcal{S}$.
$\left | S \right |$ represents the number of sentences and $\left | j: w_i^s \in d_j \right |$ denotes the number of sentences containing the word $w_i^s$.
The frequency-word set $\mathbf{W}_{freq} = \langle (w_{i}^s, r_i), ...,  (w_{j}^s, r_j) \rangle$ are obtained by sorting the output ($tf\texttt{-}idf_{(i,j)}$) from the TF-IDF algorithm, where $r_i$ denotes the frequency score.

\floatname{algorithm}{Algorithm}
\renewcommand{\algorithmicrequire}{\textbf{Input:}}
\renewcommand{\algorithmicensure}{\textbf{Output:}}
\begin{algorithm}[t]
    \caption{BiCF Mixing}
    \label{alg:codemix}
    \begin{algorithmic}[1]
        \Require $\mathcal{S}$, $\mathbf{W}_{freq}$, $\lambda_{freq}$, $\mathbf{W}_{conf}$, $\lambda_{conf}, \theta$
        \Ensure $\mathcal{T}_{mix}$
            \State $\widehat{\mathbf{W}}_{freq} \gets $ \Call{Thresh}{$\mathbf{W}_{freq}, \lambda_{freq}$}
            \State $\widehat{\mathbf{W}}_{conf} \gets $ \Call{Thresh}{$\mathbf{W}_{conf}, \lambda_{conf}$}
            \State $\widetilde{\mathbf{W}}_{sub} \gets $ \Call{Fusion}{$\widehat{\mathbf{W}}_{freq}, \widehat{\mathbf{W}}_{conf}, \theta$}
            \State $\mathcal{T}_{mix} \gets $ $\Phi $
            
            \For{$s \in \mathcal{S}$}
                \State $\widehat{s} \gets $ $s$
                \For{$w^s \in s$}
                    \If{$w^s \in \widetilde{\mathbf{W}}_{sub}$}
                        \State $w^t \gets $ \Call{Get}{$\widetilde{\mathbf{W}}_{sub}, w^s$}
                        \State $\widehat{s} \gets $ \Call{Mixing}{$\widehat{s}, w^s, w^t$}
                    \EndIf
                \EndFor
                \State $\mathcal{T}_{mix} \gets $ $\mathcal{T}_{mix} \cup \widehat{s}$
            \EndFor
            \State \Return{$\mathcal{T}_{mix}$}
    \end{algorithmic}
\end{algorithm}

Then we adopt small-scale high-quality parallel sentences (\ie, 1K), translated by skilled bilingual translators, to generate the alignments of words by using \textit{fast\_align} \cite{dyer2013simple}.
Given a few English sentences and their corresponding confidently translated sentences, the \textit{fast\_align} model uses a log-linear reparameterization of IBM Model 2 \cite{collins2011statistical} to generate a set of confidence-word pairs $\mathbf{W}_{conf} = {\langle (w_{i}^s, w_{i}^t), p_i \rangle, ...,  \langle (w_{j}^s, w_{j}^t), p_j \rangle}$ with Indonesian word and confidence score, denoted by $w_i^t$ and $p_i$, respectively.

As shown in Algorithm~\ref{alg:codemix}, after selecting the words both over the frequency threshold $\lambda_{freq}$ and the confidence threshold $\lambda_{conf}$, we then fusing words to generate the substitute words set $\mathbf{W}_{sub}$.
$Thresh$ function of line 1 and 2 in Algorithm~\ref{alg:codemix} are designed as Eq.~\ref{eq:thresh}:
\begin{equation}
\label{eq:thresh}
\widehat{\mathbf{W}} = Sort(\mathbf{W}(\cdot), \mathcal{P}(\cdot)) \odot \lambda_{(\cdot)}
\end{equation}
where $\mathbf{W}(\cdot)$ denotes frequency-word set ($\mathbf{W}_{freq}$) or confidence-word set ($\mathbf{W}_{conf}$).
$\mathcal{P}(\cdot)$ denotes frequency scores $r_i$ or confidence score $p_i$.
$\odot$ is selecting the top subset operation.
And $Fusion$ function in line 3 can be implemented as Eq.~\ref{eq:fusion}:
\begin{equation}
\label{eq:fusion}
\widetilde{\mathbf{W}} = (\widehat{\mathbf{W}}_{freq} \odot \theta) \cap (\widehat{\mathbf{W}}_{conf} \odot (1-\theta))
\end{equation}
where $\theta$ is the hyper-parameter to adjust the ratio of two branch of word sets.
Lines 4 to 13 in Algorithm~\ref{alg:codemix} illustrate the mixing procedure.
We incrementally substitute the source word  $w^s$ of a temporarily copied sentence $\widehat{s}$ with the corresponding target word $w^t$.
In this way, the mixed corpus $\mathcal{T}_{mix}$ is generated, consisting of both English words and Indonesian words.

\subsection{Cross-lingual Space Refinement}
We train and refine the initially pre-trained multilingual model (\ie, Multilingual-BERT) on the mixed corpus $\mathcal{T}_{mix}$ with annotations from the source English dataset.
This operation could update the embeddings of English words as well as the Indonesian words.
Therefore this stage allows our model to make use of English corpora and obtain a refined latent space to improve semantic representations.
The multilingual latent space can be updated with the discriminative training process as Eq.~\ref{eq:fine-tune}:
\begin{equation}
\label{eq:fine-tune}
\left\{\begin{matrix}
\Theta_{i + 1}^{l}=\Theta_{i}^{l}-\eta^{l} \cdot \nabla_{\Theta^{l}} J(\Theta)
\\ 
\eta^{i-1}=\xi \cdot \eta^{i}
\end{matrix}\right.
\end{equation}
where $\eta^l$ denotes the learning rate of the $l$-th layer.
$\Theta_{i}^{l}$ represents the parameters of the model at $l$-th layer in $i$ step.
$\nabla_{\Theta^{l}}J(\Theta)$ is the gradient of parameters $\Theta_{i}^{l}$ at $l$-th layer with regard to the model’s objective function, \ie, supervised by intent classification and slot-filling annotations in our model.

When the performance is stable on the training set (around 25 epochs in our experiments), we save the model that performs best on the validation set as the mixed refined embedding model, denoted in blue embedding space in the middle of Fig.~\ref{fig:framework}.
Then we feed fine-labeled Indonesian data into the mixed refined embedding model and transfer one more time to obtain a refined target-specific embedding model.
In this way, by utilizing the English dataset, we generate a better representation latent space for Indonesian, \ie, encoding each sentence into $\mathbb{R}^{1 \times 768}$ representation feature vector.

\subsection{Joint Decoder}
The decoder of our framework performs two tasks, \ie, intent classification and slot-filling sequence labeler, respectively.
We make use of deep bi-directional long short-term memory (Bi-LSTM) networks and a CRF layer, as shown in Eq.~\ref{eq:bilstm}, to predict the classifications for the input words \cite{chen2017improving, wang2017universal, chen2016neural, dozat2016deep}.
\begin{equation}
\label{eq:bilstm}
    h_i = [f_l(\overleftarrow{h_{i+1}}, x_i), f_r(\overrightarrow{h_{i-1}}, x_i)] \Rightarrow BiLSTM
\end{equation}
where $f_l$ is the hidden state of backward propagation and $f_r$ is the hidden state of forward feeding in BiLSTM, respectively.
And then CRF layer is appended to decode slot classes further and generate results of the framework finally.

\section{Experiments}
\subsection{Dataset and Evaluation}

We take MultiWOZ \cite{budzianowski2018multiwoz} as the English dataset and our collected ID-WOZ as the target language Indonesian dataset.
As the \textit{hospital} and \textit{police} domains in MultiWOZ contain very few dialogues (5\% of total dialogues), and only appear in the training dataset, we choose to ignore them in our experiments, following \cite{wu2019transferable}.
The \textit{train} domain is invalid in Indonesian data because it reflects the cultural difference between English and Indonesia.
Therefore, we only adopt four domains as the main experiment \textit{restaurant, hotel, taxi, attraction} shared by MultiWOZ and ID-WOZ.
Statistics of them are shown in Table~\ref{tab:domains_statistics}.
In order to suit the testing set, we have to merge the annotations of English data with Indonesian dataset, thereby abandoning a few types of labels, such as \textit{reference, choice} in MultiWOZ.
After processing, the statistics for the four domains in two datasets are reported in Table~\ref{tab:domains_statistics}.  Contrastive study of differences between our dataset and similar well-known datasets are shown in Table~\ref{t1}
All of the experiments are evaluated on the same test set from ID-WOZ (1K dialogues, 250 dialogues for each domain), which suits the local cultural background.
We use the F1 score as the evaluation metric, which is calculated by the precision rate and recall rate.

\begin{table}[]
\resizebox{0.6\textwidth}{!}{
\begin{tabular}{l|l|ccc}
\hline
\textbf{Dataset}
& \textbf{Domains} & \textbf{\# Sentences} & \textbf{\# Slots} & \textbf{\# Intent} \\ \hline

\multirow{4}{*}{\textbf{MultiWOZ}}  
& \textbf{Restaurant}   & 62, 703   & 28, 351  & 41, 177   \\
& \textbf{Hotel}        & 64, 284   & 25, 985  & 42, 434   \\
& \textbf{Taxi}         & 48, 080   & 7, 160   & 28, 976   \\
& \textbf{Attraction}   & 55, 186   & 21, 004  & 34, 053   \\ \hline

\multirow{4}{*}{\textbf{ID-WOZ}} 
& \textbf{Restaurant}   & 28, 095   & 5, 809   & 22, 312   \\
& \textbf{Hotel}        & 30, 865   & 8, 720   & 24, 694   \\
& \textbf{Taxi}         & 28, 178   & 6, 038   & 22, 168   \\
& \textbf{Attraction}   & 36, 523   & 9, 198   & 29, 513   \\ \hline

\end{tabular}}
\caption{Statistics for total number in four domains.}
\label{tab:domains_statistics}
\end{table}

\begin{table*}
\Large
\resizebox{\textwidth}{!}{
\begin{tabular}{l|rrrrrrrr}
\hline
\textbf{Dataset}    & \textbf{Twitter} & \textbf{Ubuntu} & \textbf{Sina Weibo} & \textbf{WOZ 2.0} & \textbf{Frames} & \textbf{M2M} & \textbf{MultiWOZ} & \textbf{ID-WOZ} \\ \hline
Domains & Unrestricted     & Ubuntu          & Unrestricted        & Unrestricted     & Unrestricted    & Unrestricted & 7       & 9               \\
Language & English          & English         & Chinese             & English          & English         & English      & English           & Indonesian (+En)                   \\
Total \# dialogues  & 1.3M             & 930K            & 4.5M                & 600              & 1, 369            & 1, 500         & 8, 438              & 9,189 (+1k)                      \\
Total \# tokens    & -                 &  -               & -                   & 50, 264            & 251, 867          & 121, 977       & 1, 520, 970           & 1, 551, 591                         \\
Avg. \# Turns      & 2.10             & 7.71             & 2.3                  & 7.45             & 14.60           & 9.86         & 13.68             & 13.67                         \\
Avg. \# slots      & -                &  -               &  -                   & 4                & 61              & 14           & 25                & 8.8                         \\ \hline
\end{tabular}}
\caption{Comparison of our dataset to similar well-known datasets.}
\label{t1}
\end{table*}

\subsection{Model Settings}
\label{model_setting}
There are three branches of methods to utilize English dataset and pre-trained models, \ie, \textbf{M}achine \textbf{T}ranslation based (\textbf{MT}); \textbf{M}ulti\textbf{l}ingual pre-trained embedding model with \textbf{En}glish corpus (\textbf{MLEn});
and our proposed \textbf{BiCF}.

\textbf{1) MT.}
We adopt the machine translation preprocessing method and extract word embeddings (\ie, $\mathbb{R}^{1 \times 768}$) by random initiation, pre-trained multilingual-BERT (ML-BERT) and ID-BERT.
We also take Indonesian-fastText (ID-fastText) \cite{joulin2016fasttext}, Transformer \cite{vaswani2017attention} and our pre-trained Indonesian-Word2vec (ID-Word2vec) into comparison. ($\mathbb{R}^{1 \times 300}$)

\textbf{2) MLEn.}
We adopt three pre-trained multilingual word embedding models in this baseline, namely multilingual fastText (ML-fastText) \cite{joulin2016fasttext}, multilingual Word2vec (ML-Word2vec) \cite{de2017multilingual}, and multilingual-BERT (ML-BERT) \cite{devlin2018bert}.
By extracting the embeddings of MultiWOZ and ID-WOZ, we encode each sentence into $\mathbb{R}^{1 \times 300}$, $\mathbb{R}^{1 \times 300}$ and $\mathbb{R}^{1 \times 768}$ dimensions, respectively.

\textbf{3) BiCF.}
We generate about 1.5K confident word pairs from MultiWOZ and 1K translated parallel sentences.
For our method BiCF, the training process converges after 20 epochs. 
It reaches 91.13, 87.84/ 90.17, 82.09/ 93.37, 82.98/ 89.55, 85.54 for the F1 score of intent classification and slot-filling on the MultiWOZ validation set of \textit{restaurant, hotel, taxi, attraction} domains, respectively.
And then the Indonesian training data of ID-WOZ is fed to refine the Indonesian embedding model.


\subsection{Development Experiments}
We feed 16K Indonesian sentences of ID-WOZ to each method and validate their performance on same test set of ID-WOZ.
In our implementation, five-fold cross-validation is employed to investigate the optimal parameter setting within training datasets ($learning\_rate=e^{-3}, batch\_size=64, dropout\_rate=0.1, optimizer=sgd $).
To verify the stability of the proposed method, we run the experiments five times for each set of parameter settings and compare their mean performance, reported in Table~\ref{tab:result1}.

\begin{table*}[]
\scriptsize
\resizebox{\textwidth}{!}{
\begin{tabular}{rl|rr|rr|rr|rr}
\hline
\multicolumn{2}{l|}{\multirow{2}{*}{\diagbox[width=12em,trim=l]{\textbf{Methods + Emb.}}{\textbf{Domains}}}} & \multicolumn{2}{c|}{\textbf{Restaurant}} & \multicolumn{2}{c|}{\textbf{Hotel}} & \multicolumn{2}{c|}{\textbf{Taxi}} & \multicolumn{2}{c}{\textbf{Attraction}} \\ \cline{3-10}
& & Intent & Slots & Intent & Slots & Intent & Slots & Intent & Slots \\ \hline
\textbf{MT} & \textbf{Random Init} & 85.48 & 74.36  & 82.73 & 73.49 & 89.15 & 80.22 & 89.64 & 86.26 \\
\textbf{MT} & \textbf{ID-fastText} & 86.03 & 75.27 & 83.17 & 74.03 & 89.82 & 80.28 & 90.02 & 86.88 \\
\textbf{MT} & \textbf{ID-Word2vec} & 88.22 & 76.70 & 86.33 & 74.11 & 89.91 & 81.81 & 91.55 & 86.90 \\
\textbf{MT} & \textbf{Transformer} & 90.13 & 79.91 & 91.89 & 74.27 & 90.25 & 82.11 & 92.85 & 87.16 \\
\textbf{MT} & \textbf{ML-BERT} & 91.63 & 79.22 & 92.52 & 73.83 & 91.20 & 82.34 & 93.77 & 87.31 \\
\textbf{MT}& \textbf{ID-BERT} & \textit{92.37} & \textit{81.88} & \textit{93.78} & \textit{75.79} & \textit{91.76} & \textit{83.59} & \textit{94.07} & \textit{89.63} \\ 
\hline
\textbf{MLEn} & \textbf{ML-fastText} & 86.00 & 76.11 & 83.10 & 74.91 & 89.22 & 80.88 & 90.31 & 86.93 \\
\textbf{MLEn} & \textbf{ML-Word2vec} & 88.22 & 77.70 & 86.33 & 74.11 & 89.91 & 81.81 & 91.55 & 86.90 \\
\textbf{MLEn} & \textbf{ML-BERT} & \textit{90.42} & \textit{79.79} & \textit{92.01} & \textit{74.28} & \textit{90.47} & \textit{82.91} & \textit{93.18} & \textit{87.77} \\
\hline
\textbf{BiCF} & \textbf{ML-fastText} & 86.21 & 76.16 & 83.31 & 75.01 & 90.24 & 82.58 & 90.84 & 87.23 \\
\textbf{BiCF} & \textbf{ID-fastText} & 87.08 & 76.34 & 84.21 & 75.79 & 90.83 & 82.920 & 91.52 & 87.67 \\
\textbf{BiCF} & \textbf{ML-Word2vec} & 88.80 & 77.91 & 87.12 & 74.24 & 90.01 & 82.87 & 91.58 & 87.03 \\
\textbf{BiCF} & \textbf{ID-Word2vec} & 88.92 & 78.84 & 88.52 & 74.35 & 90.31 & 83.15 & 91.82 & 87.49 \\
\textbf{BiCF} & \textbf{ML-BERT} & \textit{92.92}$^{\dagger}$ & \textit{82.84}$^{\dagger}$ & \textit{94.30}$^{\dagger}$ & \textit{76.95}$^{\dagger}$ & \textit{92.23}$^{\dagger}$ & \textit{90.45}$^{\dagger}$ & \textit{94.80}$^{\dagger}$ & \textit{90.44}$^{\dagger}$ \\
\textbf{BiCF} & \textbf{ID-BERT} & \textbf{93.02}$^{\dagger}$ & \textbf{82.91}$^{\dagger}$ & \textbf{94.73}$^{\dagger}$ & \textbf{77.15}$^{\dagger}$ & \textbf{92.73}$^{\dagger}$ & \textbf{91.03}$^{\dagger}$ & \textbf{94.88}$^{\dagger}$ & \textbf{90.74}$^{\dagger}$ \\
\hline
\end{tabular}
}
\caption{Experimental comparison on ID-WOZ dataset. (``${\dagger}$'' denotes the significance testing, $p$-$value < 0.05$.)}
\label{tab:result1}
\end{table*}

\begin{figure*}
    \centering
    \includegraphics[width=\linewidth]{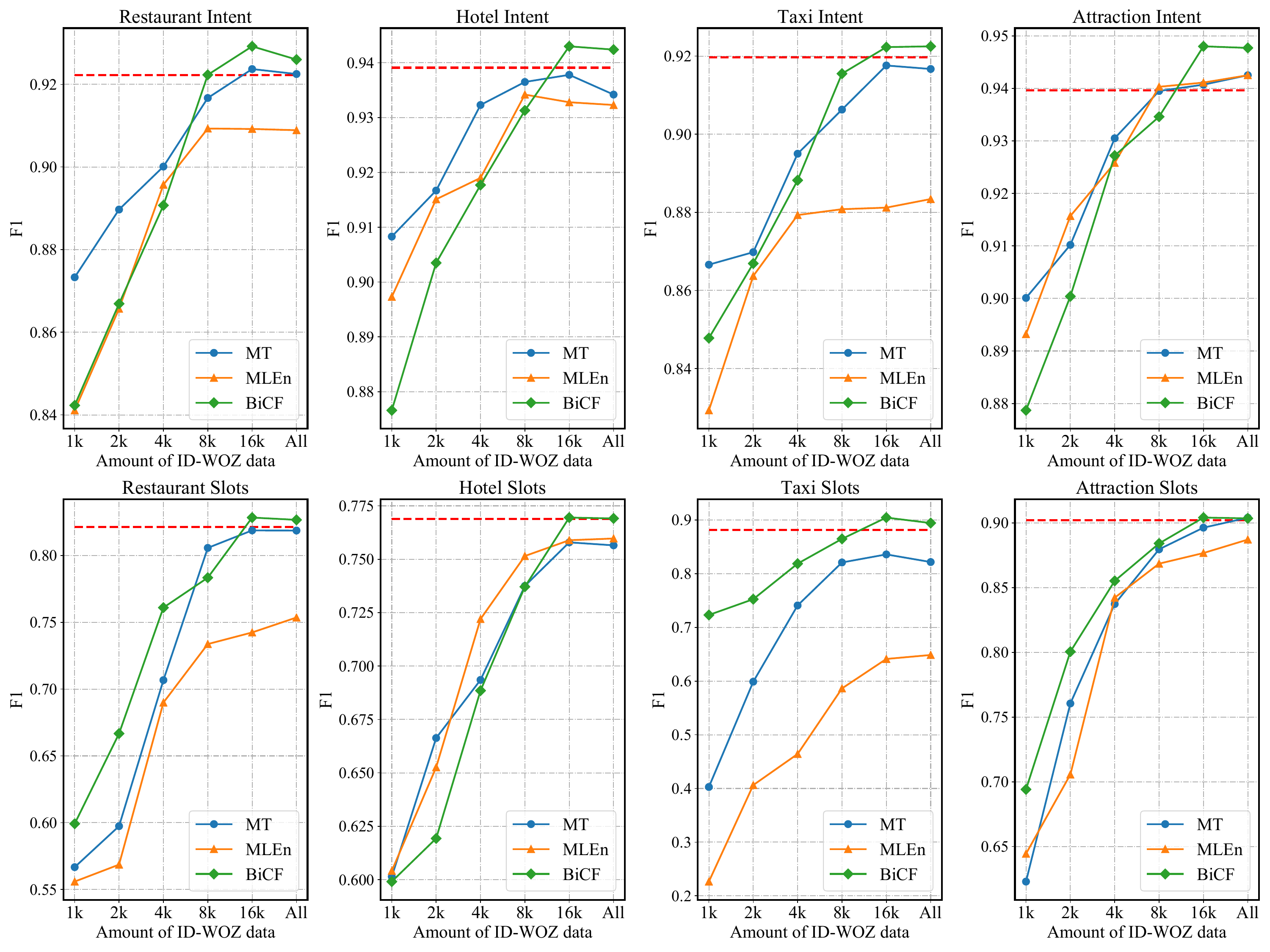}
    \caption{\label{fig:k_parameters_line} The comparison of different methods on four domains.}
\end{figure*}

\begin{figure*}
    \centering
    \includegraphics[width=\linewidth]{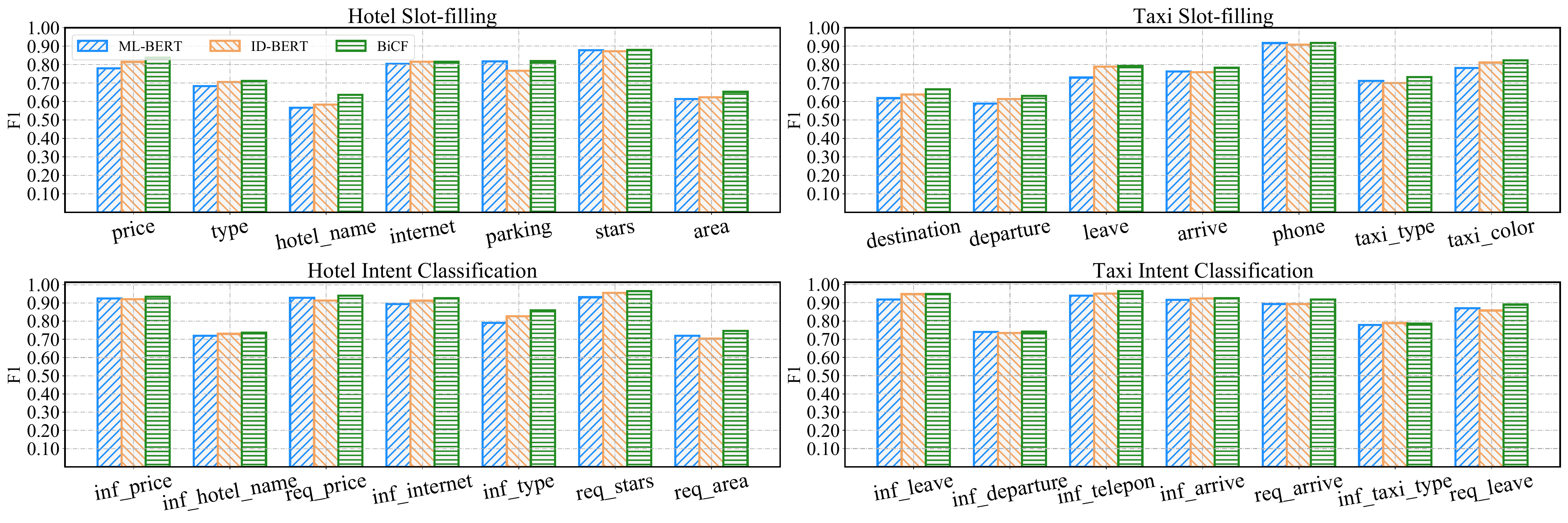}
    \caption{\label{fig:k_parameters_his} The comparison of different methods on four domains.}
\end{figure*}

We also conduct a series of experiments by feeding batches of annotated Indonesian data (\ie, 1K sentences, 2K sentences, 4K sentences, ..., full-scale).
We pick the results of \textit{restaurant}, \textit{hotel}, \textit{taxi}, and \textit{attraction} domains in Fig.~\ref{fig:k_parameters_line} and Fig.~\ref{fig:k_parameters_his}, as they are widely usable domains and have the most scale of dialogue data and annotations both in MultiWOZ and ID-WOZ.
The entire annotated dataset, experiment results and codes are detailedly reported in Table~\ref{tab:result2} and Code~\ref{alg:codemix}.
Also, we conduct a comparison experiment for Multilingual-BERT (ML-BERT) and ID-BERT on all domains of full-scale ID-WOZ, as reported in Table~\ref{tab:bert-comparison}.

\begin{table*}[]
\scriptsize
\resizebox{\textwidth}{!}{
\begin{tabular}{c|l|cc|cc|cc|cc}
\hline
\multirow{2}{*}{\textbf{Methods}} & \multirow{2}{*}{\textbf{ID-WOZ}} & \multicolumn{2}{c|}{\textbf{Restaurant}} & \multicolumn{2}{c|}{\textbf{Hotel}} & \multicolumn{2}{c|}{\textbf{Taxi}} & \multicolumn{2}{c}{\textbf{Attraction}} \\ \cline{3-10}
& & Intent & Slots & Intent & Slots & Intent & Slots & Intent & Slots \\ \hline
\multirow{6}{*}{\begin{tabular}[c]{@{}c@{}}\textbf{MT} \textbf{(ID-BERT)}\end{tabular}} 
& \textbf{ID-WOZ-1000} & 87.33 & 56.67 & 90.83 & 60.14 & 86.66 & 40.28 & 90.01 & 62.29 \\
& \textbf{ID-WOZ-2000} & 88.97 & 59.74 & 91.67 & 66.63 & 86.98 & 59.88 & 91.02 & 76.05 \\
& \textbf{ID-WOZ-4000} & 90.01 & 70.67 & 93.23 & 69.35 & 89.50 & 74.09 & 93.05 & 83.73 \\
& \textbf{ID-WOZ-8000} & 91.67 & 80.57 & 93.65 & 73.75 & 90.63 & 82.09 & 93.95 & 87.96 \\
& \textbf{ID-WOZ-16000} & 92.37 & 81.88 & 93.78 & 75.79 & 91.76 & 83.59 & 94.07 & 89.63 \\
& \textbf{ID-WOZ-All}  & 92.25 & 81.87 & 93.42 & 75.65 & 91.67 & 82.17 & 94.25 & 90.40 \\ \hline
\multirow{6}{*}{\begin{tabular}[c]{@{}c@{}}\textbf{MLEn}\\ \textbf{(ML-BERT)}\end{tabular}} 
& \textbf{ID-WOZ-1000} & 84.11 & 55.59 & 89.73 & 60.41 & 82.93 & 22.66 & 89.32 & 64.44 \\
& \textbf{ID-WOZ-2000} & 86.57 & 56.86 & 91.51 & 65.26 & 86.37 & 40.63 & 91.57 & 70.56 \\
& \textbf{ID-WOZ-4000} & 89.57 & 68.99 & 91.90 & 72.20 & 87.93 & 46.42 & 92.58 & 84.22 \\
& \textbf{ID-WOZ-8000} & 90.93 & 73.37 & 93.42 & 75.15 & 88.08 & 58.63 & 94.03 & 86.85 \\
& \textbf{ID-WOZ-16000} & 90.92 & 74.24 & 93.28 & 75.89 & 88.12 & 64.12 & 94.11 & 87.67 \\
& \textbf{ID-WOZ-All} & 90.89 & 75.36 & 93.23 & 75.97 & 88.34 & 64.86 & 94.25 & 88.71 \\ \hline
\multirow{6}{*}{\textbf{\begin{tabular}[c]{@{}c@{}} BiCF \\ (ML-BERT) \end{tabular}}}
& \textbf{ID-WOZ-1000} & 84.23 & 59.92 & 87.66 & 59.81 & 84.78 & 72.31 & 87.87 & 69.41 \\
& \textbf{ID-WOZ-2000} & 86.69 & 66.67 & 90.35 & 61.93 & 86.69 & 75.25 & 90.04 & 80.05 \\
& \textbf{ID-WOZ-4000} & 89.07 & 76.10 & 91.77 & 68.85 & 88.82 & 81.87 & 92.72 & 85.52 \\
& \textbf{ID-WOZ-8000} & 92.23 & 78.34 & 93.13 & 73.71 & 91.55 & 86.48 & 93.46 & 88.41 \\
& \textbf{ID-WOZ-16000} & \textbf{92.92} & \textbf{82.84} & \textbf{94.30} & \textbf{76.95} & 92.23 & \textbf{90.45} & \textbf{94.80} & 90.44 \\
& \textbf{ID-WOZ-All} & 92.60 & 82.67 & 94.24 & 76.91 & \textbf{92.25} & 89.43 & 94.77 & \textbf{90.45} \\ \hline
\textbf{ID-BERT} & \textbf{ID-WOZ-All} & 92.22 & 82.14 & 93.91 & 76.88 & 91.97 & 88.13 & 93.96 & 90.20 \\ \hline
\end{tabular}
}
\caption{Performance comparison of different methods on the selected MultiWOZ and ID-WOZ with different amounts of feeding ID-WOZ data.}
\label{tab:result2}
\end{table*}

\begin{table}[]
\begin{tabular}{l|cccc}
\hline
\multirow{2}{*}{\textbf{Domains}} & \multicolumn{2}{c}{\textbf{ML-BERT}} & \multicolumn{2}{c}{\textbf{ID-BERT}} \\
 & Intent & Slots & Intent & Slots \\ \hline
\textbf{Restaurant} & 91.07 & 77.68 & 92.22 & 82.14 \\
\textbf{Hotel} & 92.78 & 74.91 & 93.91 & 76.88 \\
\textbf{Taxi} & 90.84 & 82.91 & 91.97 & 88.13 \\
\textbf{Attraction} & 93.25 & 88.04 & 93.96 & 90.20 \\
\textbf{Plane} & 91.36 & 92.77 & 93.42 & 93.11 \\
\textbf{Police} & 90.02 & 88.89 & 92.78 & 90.07 \\
\textbf{Movie} & 90.57 & 86.14 & 91.76 & 87.98 \\
\textbf{Hospital} & 92.64 & 84.15 & 93.85 & 86.09 \\
\textbf{Wear} & 90.77 & 87.02 & 91.80 & 88.34 \\ \hline
\end{tabular}
\caption{Experimental comparison of ML-BERT and ID-BERT on full-scale ID-WOZ.}
\label{tab:bert-comparison}
\vspace{-0.4cm}
\end{table}

\subsection{Results Analysis}
The results of the method in Section~\ref{model_setting} are shown in Table~\ref{tab:result2}, with English data of MultiWOZ and 16k Indonesian data of ID-WOZ.
The method of machine translation based methods (MT + ML-BERT/ ID-BERT) surpass multilingual model with English data (MLEn + ML-BERT) on the intent classification task, outperforming by about 1.21\%, 1.95\%; 0.51\%, 1.77\%; 0.73\%, 1.29\% and 0.59\%, 0.89\% on F1 score for \textit{restaurant, hotel, taxi, attraction}, respectively.
The main reason is that the machine translation methods enjoy much more Indonesian sentences with corresponding intention labels.
However, on the slot-filling task, the machine translation methods are weaker.
Because the machine translation methods suffer from invalid or mismatching labels after translation.
Overall, our proposed framework (BiCF + ML-BERT / ID-BERT) performs better than others in both tasks, as we are capable of utilizing the English intention labels and correct slot-filling annotations effectively.
And from Table~\ref{tab:bert-comparison}, we can see that ID-BERT outperforms ML-BERT across all domains, demonstrating Indonesian-specific word-embedding model (ID-BERT) is capable of representing more information and semantic knowledge than the general multilingual model (ML-BERT) in all of domains.

\subsection{Effectiveness of Using ID-WOZ}
The statistics line chart is shown in Fig.~\ref{fig:k_parameters_line}, where the four upmost sub-graphs denote intent classification, the four downmost sub-graphs denote slot-filling and the red line is the performance of ID-BERT baseline.
The detailed results and all of the line charts of rest domains are in Figure~\ref{fig:k_parameters_his}.

\begin{figure}
    \centering
    \includegraphics[width=0.6\linewidth]{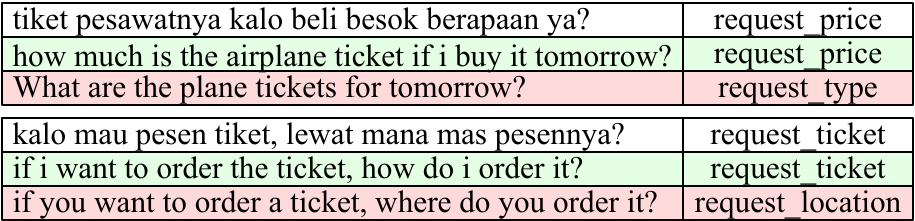}
    \caption{\label{fig:eg_trans_prob} Illustration of the mistakes from machine translation. The green sentence is true mean, and the red is the result of machine translation. These two examples show that a tiny mistake that happened during translation may cause complete misunderstanding.}
\end{figure}

\textbf{1).} 
\textbf{MT} methods rely heavily on the quality of translation.
We conduct the BLEU \cite{papineni2002bleu} test for the entire MultiWOZ, and the performance of translation is 28.46 (BLEU-5) on 30k sentences.
However, during the translation of dialogue, one incorrect word would cause misunderstanding.
Several examples are shown in Fig.~\ref{fig:eg_trans_prob}.
Different sentences in English may be translated from the same source sentence in Indonesian.
In the first case, the true meaning is requesting ``\textit{how much}'' but the model may misunderstand the customer's intent into requesting the type of plane ticket.
And in the second, the customer is wondering ``how'' to order a ticket, but the translator gives the result that the customer's request is``\textit{request location}''.
Based on Fig.~\ref{fig:k_parameters_line} and Fig.~\ref{fig:k_parameters_his}, when the scale of ID-WOZ is negligible, the machine translation has large advantage on intent classification but performs badly on slot-filling.
The reason is that the MT method has the ability to adjust or reset the grammar and syntactic structure to the target language, whose characteristic leads to the bad consequences that make the English slots labels dislocated, invalid and wrong.



\textbf{2).}
\textbf{MLEn} methods only learn semantic information from English data in the beginning, which causes low accuracy on intent classification than others.
When feeding this model with ID-WOZ, it has weakness coming from the English data because the large-scale English data shrinks the feeding ID-WOZ data.
This method has strength on slot-filling when the comparison is under small scale of ID-WOZ.
Because labels of slot-filling in the English data are accurate and complete.
But the performance does not improve when more ID-WOZ data is further used, which shows ML-BERT has a limitation on reaching higher performance.
Overall, this method is not recommended for building stable low-resource language dialogue understanding models even with gold annotated data.

\begin{figure*}
    \centering
    \includegraphics[width=\linewidth]{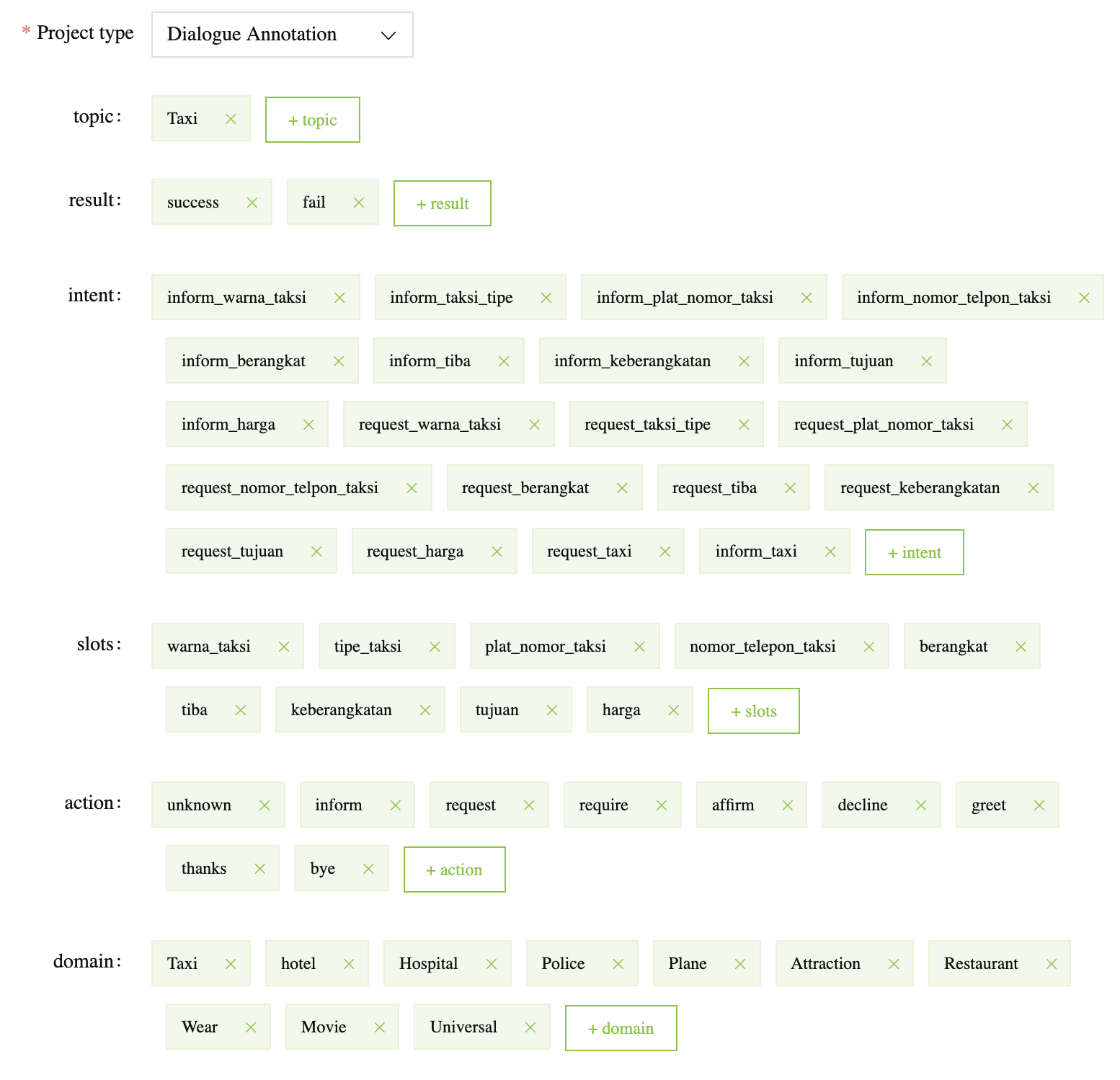}
    \caption{An example of editing template interface.}
    \label{f_abs}
\end{figure*}

\begin{figure*}
    \centering
    \includegraphics[width=\linewidth]{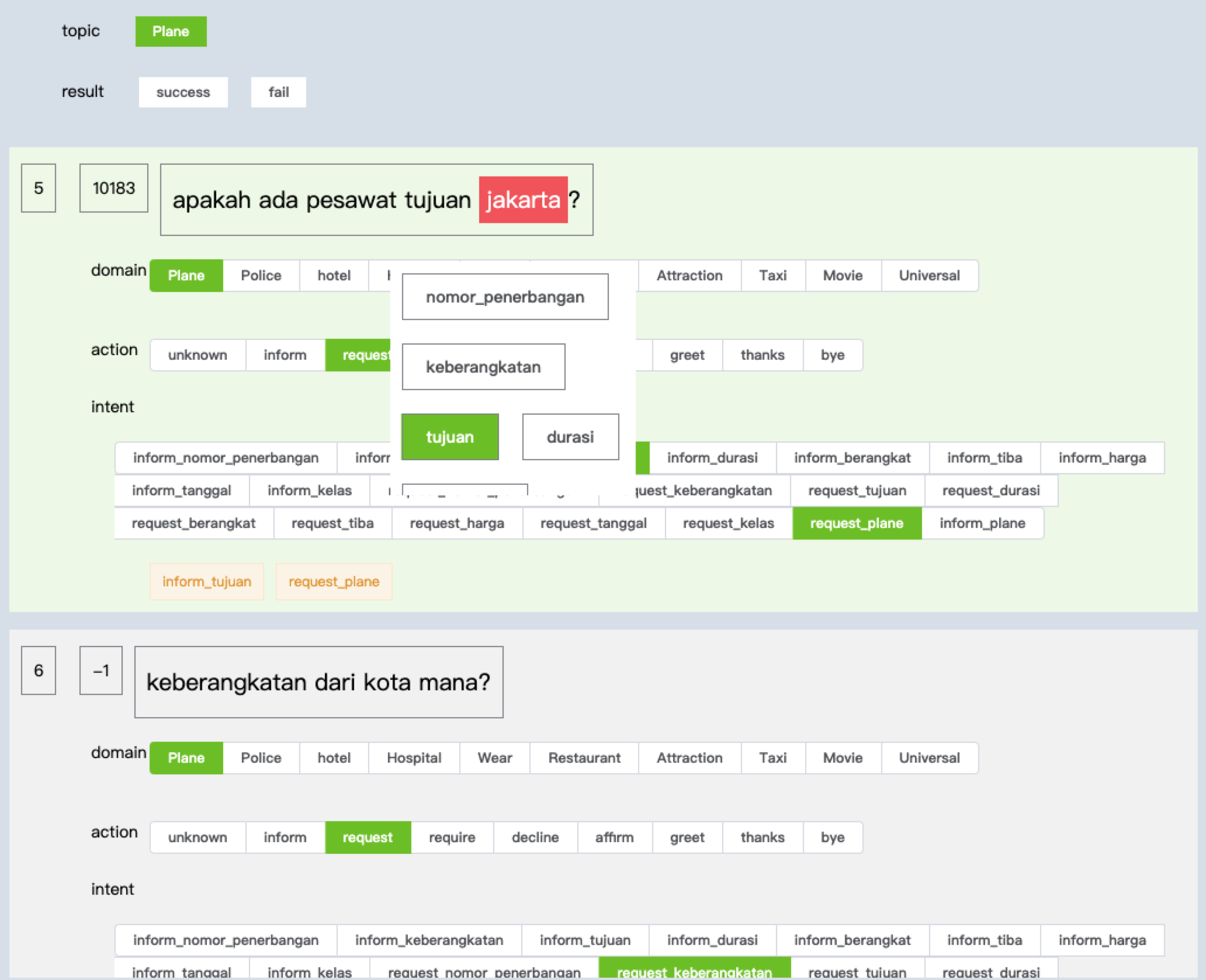}
    \caption{An example of the annotation procedure. For domain/action/intent classification, the annotator could click these multiple labels, defined before for each domain. As for slot-filling, our platform provides a fashion approach: click and underscore the content and select its slot type, pop-out automatically when any words are picked.}
    \label{f_anno}
\end{figure*}

\begin{figure*}
    \centering
    \includegraphics[width=\linewidth]{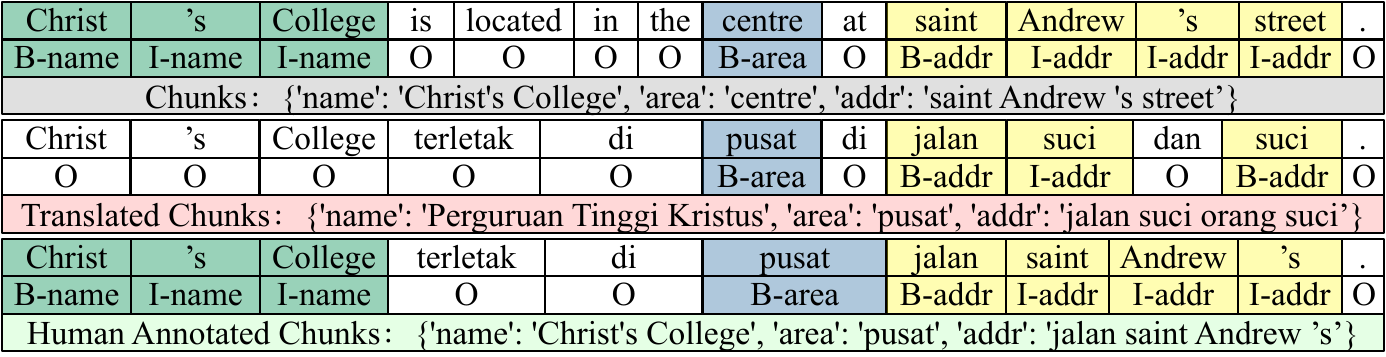}
    \caption{\label{fig:k_parameters_mt} Illustrate the situation that annotations getting invalid in the machine translation.}
\end{figure*}

\textbf{3).} 
\textbf{BiCF} does not outperform machine translation when the scale of fed Indonesian data is negligible on the intent classification.
When the scale of ID-WOZ data gets larger, its strength of BiCF becomes more obvious.
It starts to outperform significantly better than the other methods while the Indonesian data grows.
It is capable of avoiding misunderstanding caused by translation and mitigating the shrink effect of the English corpus, which makes it achieve the best performance and even better than the baseline ID-BERT, when the ID-WOZ data reaches around 16k for \textit{restaurant, hotel, taxi, attraction} domains on the intent classification, \ie, 92.92\%, 94.30\%, 92.23\%, 94.80\% on F1 score, respectively.
This method outperforms other methods on slot-filling when the ID-WOZ data fed is negligible. 
Not only it makes use of correct slot-filling annotations from the English dataset, but it can also reduce the bad effects of large-scale English corpus.
The accuracy reaches 82.84\%, 76.95\%, 90.45\%, 90.44\% on F1 score for \textit{restaurant, hotel, taxi, attraction} on the slot-filling, respectively.
Fig.~\ref{fig:k_parameters_his} reports the results of three methods trained by 16K of ID-WOZ. 
It shows that the cross-lingual method performs better than others when the slots need more words to describe.

\section{Conclusion and Future Work}
We empirically investigated how to build the low-resource language dialogue understanding models with the English dataset from scratch.
Directly translating from English to Indonesian or simply utilizing the multilingual pre-trained model could not perform well.
Instead, our framework BiCF enjoys the enriched and accurate English dataset, performs effectively and obtains reliable results.
We further release a large Indonesian dialogue dataset and an ID-BERT model for future research.

\section{Acknowledgments}

This research is supported by the
Nature Scientific Foundation of Heilongjiang Province (YQ2021F006)

\bibliographystyle{ACM-Reference-Format}
\bibliography{cites}

\end{document}